\documentclass[runningheads]{llncs}
\usepackage[T1]{fontenc}
\usepackage{amsfonts}
\usepackage{graphicx}
\usepackage{hyperref}
\usepackage[numbers]{natbib}
\setcitestyle{numbers,square}
\usepackage{algorithm}
\usepackage{amsmath}
\usepackage{booktabs}
\begin{document}
\title{Supervised Contrastive Frame Aggregation for Video Representation Learning}

\author{Shaif Chowdhury\inst{1} \and
Mushfika Rahman\inst{2} \and
Greg Hamerly\inst{3}}
\institute{Texas A\&M University Kingsville, TX, USA, \email{shaif.chowdhury@tamuk.edu}, \url{shaifchowdhury.com}\and
Baylor University, TX, USA,
\email{Greg\_Hamerly@baylor.edu}\\ \and
Baylor University, TX, USA\\
\email{mushfika\_rahman1@baylor.edue}}
\maketitle              
\begin{abstract}

We propose a novel supervised contrastive learning framework for video representation learning that can leverage temporally global context. First, we introduce a video-to-image aggregation strategy, where several frames from each video are spatially arranged into a single input image. This design allows us to use pre-trained off-the-shelf CNN backbones (e.g. ResNet-50) and avoids the computational overhead of more complex video transformer models\cite{chowdhury2024efficient}.   

Second, we design a contrastive learning objective that directly compares the pairwise projections generated by the model. Positive pairs are defined as projections from videos with the same label, whereas all other projections are contrasted as negatives. We create multiple natural “views” of the same video using different temporal frame samplings of the same underlying video. Rather than relying on augmentations, these frame-level changes yield diverse positive samples with global context, mitigating over-fitting and creating good discriminative representations.

Empirical results in Penn Action and HMDB51 demonstrate the effectiveness of our approach, surpassing state-of-the-art methods in classification accuracy while demanding fewer computational resources. The proposed \textbf{Supervised Contrastive Frame Aggregation (SCFA)} strategy can learn quality video representations in both supervised and self-supervised settings, thus providing an efficient solution for video-based tasks like classification, captioning, etc. In particular, our Supervised Contrastive Frame Aggregation (SCFA) method achieves 76\% classification accuracy in Penn Action over 43\% with the ViVIT model. We also get 48\% accuracy in the video classification on HMDB51, which is significantly superior to the ViVIT model that achieves 37\% accuracy.

\keywords{Video Recognition  \and Contrastive Learning \and Aggregation.}
\end{abstract}

\section{\textbf{Introduction}}
\label{Introduction}

Deep learning has revolutionized artificial intelligence by enabling hierarchical feature learning from raw data\cite{devlin2018bert,chowdhury2025deep,arnab2021vivit,quevedo2023fairness,abbasi2025leveraging,hamara2025learning,sanjel2025nlp,hamara2025learning,rahman2025ai2ds,acharya2025high}. It has become a cornerstone for video analysis, enabling applications ranging from video classification\cite{chowdhury2023video} to captioning and video generation \cite{chowdhury2024acfed,arnab2021vivit}. Despite this success, video representation learning remains particularly challenging due to the temporal dimension\cite{chowdhury2022recognition}, which often introduces a high computational overhead in both training and inference \cite{sharma2021video}. Existing approaches typically adopt architectures like 3D convocational neural networks (CNNs)\cite{rasham2024glioseg} or video transformers\cite{arnab2021vivit} to model spatio-temporal features, yet these methods tend to be resource intensive \cite{schiappa2023self}. In light of these challenges, we propose a Supervised Contrastive Frame Aggregation framework to improve both efficiency and accuracy in learning robust video representations.

\subsection{\textbf{Self-supervised Learning}}

Self-supervised learning \cite{krishnan2022self} is widely used in scenarios where labeled data are scarce, but a large amount of unlabeled data is available. For example, when a data set is collected over a long period of time, camera parameters and environmental conditions can change substantially\cite{sigurdsson2018charades}, causing the test data to differ from the training data \cite{xu2019self}. Although humans can readily adapt to such variations, deep learning models often experience degraded performance. A man who sings on a baseball field (before the start of the game) might be mistaken for a baseball player, by a model trained in a supervised way. This is mainly because the training data set generally does not have any examples of a man singing on a basketball court. Humans would generally not make such mistakes and can generally learn a task from a few examples. This happens acquired background knowledge that allows us to learn new concepts from relatively few examples. 

Common approaches to self-supervised learning include pretext task learning using pseudo-labeling, encoder-based learning, and contrastive learning \cite{balestriero2023cookbook}. In biological species detection, pseudo-labeling has proven effective for tasks such as leaf disease classification~\cite{qin2016identification}, invasive species detection~\cite{kim2024predicting}, and underwater fish segmentation~\cite{saleh2024track}. However, the success of pretext-based representation learning is heavily dependent on the alignment between pretext and downstream tasks~\cite{wang2023tree,rani2023self,metaxassimplifying}. Tasks like rotation prediction or jigsaw solving do not generalize well when the downstream objective (e.g., species segmentation) and requires more spatially rich features~\cite{wang2023tree}.

Contrastive learning~\cite{tian2020makes} is one of the major advances in self-supervised representation learning. In contrastive learning, a model generally learns to create a representation by comparing and contrasting an augmented version of an image. The self-supervised contrastive loss function enforces representations of augmented versions of the same image to be closer, while pushing apart representations of different images. This algorithm generates a supervised signal through data augmentation.

\subsection{\textbf{Motivation}}

For large-scale and accurate video recognition models, it is important to be able to learn without labels. For image recognition, contrastive learning has emerged as a solution to this. Although contrastive learning has made a lot of progress in images and multimodal models, it still remains underexplored in the video space \cite{kumar2022contrastive}. A lot of contrastive learning methods are based on augmentations that don't apply directly to video models. To deal with this, it is natural to use temporal signals from a video for contrastive learning \cite{le2020contrastive}. Video models often also require massive amounts of compute for pre-training, especially for temporally long video sequences. In this paper, we try to come up with a contrastive learning method that can address these factors.

\begin{figure*}[h]
    \centering
    \includegraphics[width=.6\textwidth, height = .25\textwidth]{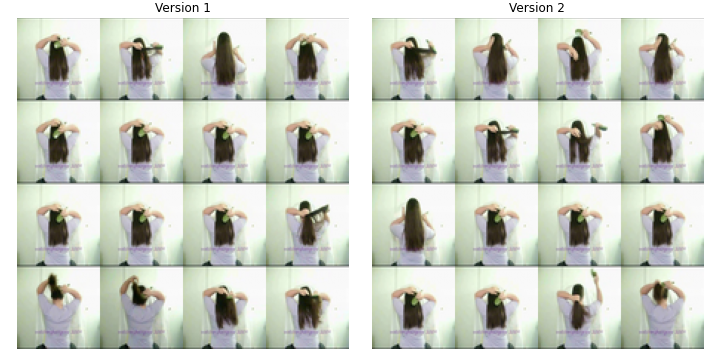}
    \caption{Two different sampled versions of the same video aggregated into a single image, highlighting the temporal diversity captured by our framework.}
    \label{fig:version_comparison}
\end{figure*}

\subsection{\textbf{Supervised Contrastive Frame Aggregation}}

Our proposed framework, \textbf{Supervised Contrastive Frame Aggregation}, introduces a novel approach to video representation learning by leveraging temporal context through frame aggregation and contrastive learning. The key idea is to spatially arrange multiple frames from a video into a single image, enabling the use of pre-trained image-based CNN backbones (e.g. ResNet-50)\cite{koonce2021resnet} without the need for computationally expensive 3D convolutions or video transformers. This design allows us to efficiently capture the global temporal context while maintaining computational efficiency.

The framework operates in two stages:

\begin{itemize}
    \item \textbf{Video-to-Image Aggregation}: For each video, we sample a set of frames and arrange them into a grid-like structure, creating a single composite image. The frames are arranged from left to right and from top to bottom in a single image. This aggregation preserves the temporal information of the video while transforming it into a format compatible with standard image-based models.
    \item \textbf{Contrastive Learning with Temporal Signals}: We define positive pairs as projections from videos with the same label, while negative pairs are formed from videos with different labels. To create diverse positive samples, we generate multiple "views" of the same video by sampling different sets of frames. This approach leverages the temporal signals naturally present in the video, avoiding the need for artificial augmentations and ensuring robust representations.
\end{itemize}

By combining these strategies, SCFA achieves state-of-the-art performance in video classification tasks while significantly reducing computational overhead. The use of pre-trained image models further enhances its efficiency, making it a practical solution for large-scale video analysis.

Our work makes the following key contributions:
\begin{itemize}
    \item \textbf{Efficient Video-to-Image Aggregation}: We propose a frame aggregation strategy as shown in \ref{fig:version_comparison}.

    \item  \textbf{Model Architecture}: We adopt a dual input architecture where two aggregated images go through a shared CNN backbone(such as ResNet50) and projection layers creating two different embeddings. These projections are passed to our contrastive learning loss function along with their labels, which optimizes them to be closer to the embedding space. This design works seamlessly with our contrastive objective, ensuring that the learned representations preserve both temporal and spatial information from the videos.
    
    \item \textbf{Flexibility and Generalizability}: The proposed framework can be applied in both supervised and self-supervised settings, making it versatile for various video-based tasks such as classification, captioning, and retrieval.
    
    \item \textbf{Empirical Validation}: We demonstrate the effectiveness of our approach on benchmark data sets (Penn Action and HMDB51), achieving state-of-the-art classification accuracy with significantly lower computational requirements compared to existing methods.
    
\end{itemize}

\section{\textbf{Related Work}}
\label{Related Work}

Various self-supervised learning methods were proposed to address the issue of supervised learning methods that require expensive and time-consuming human annotations. One of the popular solutions is the introduction of pretext tasks such as colorizing images ~\cite{zhang2016colorfulimagecolorization}, image rotation ~\cite{gidaris2018unsupervisedrepresentationlearningpredicting}, image jigsaw puzzle ~\cite{noroozi2017unsupervisedlearningvisualrepresentations}, and many more to learn the visual features of unlabeled large-scale data. Additionally, some self-supervised learning methods rely on pseudo-labels to learn visual features\cite{choudhury2017vehicle}. Researchers have indicated that augmentations as pretext tasks allow the model to learn robust and useful features without labeling ~\cite{qian2021spatiotemporal}. Furthermore, self-supervised instance-level discrimination methods played a significant role in achieving the goal of representation learning. In particular, contrastive learning takes advantage of data augmentations and instance discrimination to learn meaningful visual representations. The underlying idea is to maximize the similarity between augmented views of the same image (positive pairs) and minimize the similarity between different images (negative pairs) ~\cite{chen2020simple}. For example, methods such as SimCLR ~\cite{chen2020simple,tisha2024contrastive} and MoCo ~\cite{he2020momentum} maximize the agreement between two augmented versions of the same image from the rest of the samples. The methods use strong augmentation techniques such as random cropping, color jittering, and Gaussian blur. However, the contrastive loss requires more challenging negative examples. Methods such as BYOL ~\cite{grill2020bootstrap}, SimSiam ~\cite{chen2021exploring}, and Barlow Twin ~\cite{zbontar2021barlow} modify the loss function without the need for negative pairs. However, the potential issue in eliminating negative pairs is that all representations can lack diversity. The method uses stop-gradients and momentum encoders to avoid trivial solutions, making the model computationally expensive ~\cite{chen2021exploring, grill2020bootstrap}. supervised learning environments heavily rely on labels to separate classes; thus, incorporating labels in self-supervised settings can significantly improve, introducing the concept of supervised contrastive learning ~\cite{khosla2020supervised}.

\subsection{\textbf{Self Supervised Learning in Video}}

Due to the extra temporal dimension, extending well-established self-supervised learning to video challenges arises. An apparent solution is to sample two clips from the same video and consider them a positive pair, whereas clips from different videos are considered negatives. The concept is straightforward; the instance is recognized from a specific video instead of a single static image. Augmentation plays a pivotal role ~\cite{qian2021spatiotemporal}. Since data augmentation now spans time and space, naively applying augmentation may harm the representation learning process ~\cite{qian2021spatiotemporal}. Methods need to learn meaningful representation, which can aid in useful downstream tasks, such as action recognition. In action recognition, the model needs to recognize the appearance of the object and how it changes ~\cite{jhuang2013towards}. Researchers have proposed several multi-stage training processes and architectures ~\cite{li2023spatial,qian2021spatiotemporal,dave2022tclr}; however, it is an area of active exploration.

Our proposed solution for video-to-image aggregation considers temporal and spatial features capture global context and leverages a simple architecture. Instead of relying heavily on augmentation for positive-negative samples, we incorporate labels. Thus, our proposed solution becomes less computationally expensive and scalable, producing a good representation in the downstream tasks.

\section{\textbf{Method}}
\label{Method}

\begin{figure}[h]
    \centering
    \includegraphics[width=.3\textwidth, height = .45 \textwidth]{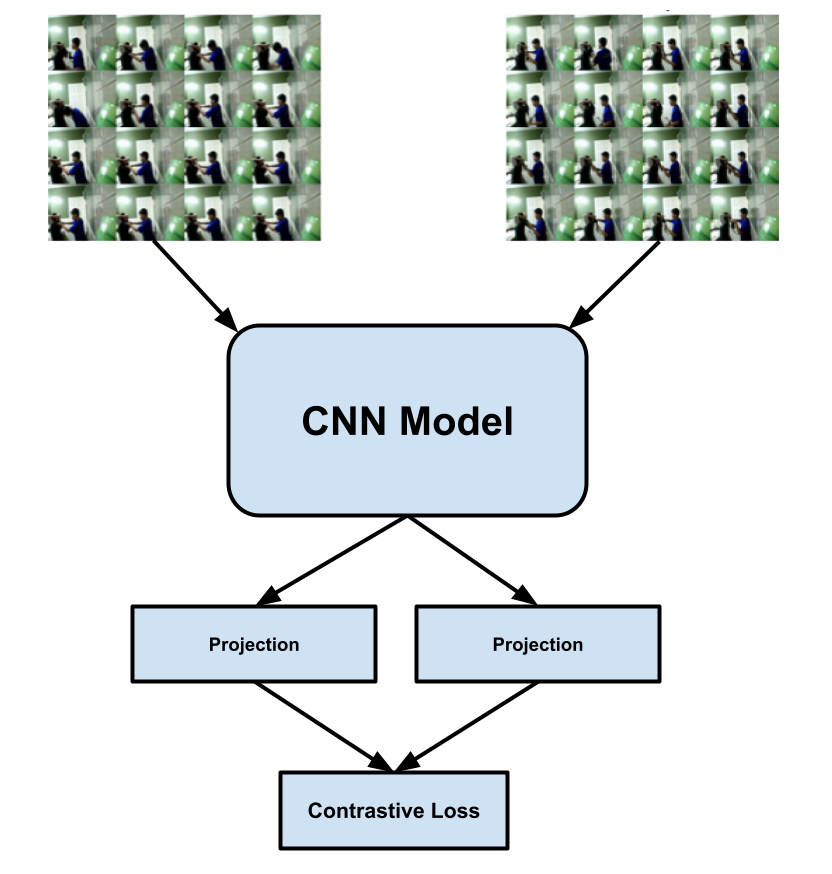}
    \caption{Overview of the Supervised Contrastive Frame Aggregation (SCFA) architecture. Two temporally sampled views of the same video are aggregated into images and passed through a shared CNN model and projection head. The resulting embeddings are then optimized using a supervised contrastive loss.}
    \label{fig:scfa_architecture}
\end{figure}

Before diving into our framework, we visit contrastive learning and self-supervised learning in the area of images. Self-supervised learning in the image domain is often based on creating positive pairs through image-based data augmentation like crop, flip, rotate, blur, etc. The core idea behind contrastive learning is that an encoder is trained to create representations from images. These representations are closer to each other for similar images or positive pairs and farther from each other for negative pairs. In general, obtains supervised signals from positive pairs created through augmented versions. Supervised contrastive learning improves on this by including the label information along with the images. That means that there are more positive examples based on the similarity of labels with augmented versions. In extending contrastive learning from images to videos, one key adaptation involves leveraging the temporal dimension to create diverse positive pairs. For video contrastive learning, this diversity of positive views can be obtained by randomly sampling different frames from a video. A frame aggregation method along with that allows the model to also take advantage of pre-trained CNN models and extract good spatial features in an efficient way.  A diagram of the method is given in Figure \ref{fig:scfa_architecture}.

\subsection{\textbf{Problem Formulation:}} Formally, given a video $V^{a}$ which contains n frames given by \{$v^{a}_{1}$,$v^{a}_{2}$, $v^{a}_{3}$, .... $v^{a}_{n}$  \}, the goal is to train an encoder network that can create a representations \{$r^{a}_{1}$,$r^{a}_{2}$, $r^{a}_{3}$, .... $v^{a}_{n}$  \}. A typical contrastive learning framework has four different components described in the following : 

\begin{itemize}

    \item \textbf{Augmentation Module} This module generally creates positive pairs from an image through some form of augmentation.

    \item \textbf{Encoder} The encoder network would take the images along with their augmented version and create a representation from them. For a video $V^{a}$, the encoder would create the representation $r^{a}$, where $r^{a}$ = Enc ( $V^{a}$ ).

    \item  \textbf{Projection} A linear projection layer maps $r^{a}$ using. This is generally a Dense layer f(z), where contrastive loss is applied, $z^{a}$ = f($r^{a}$)

    \item  \textbf{Contrastive Loss Function} For self-supervised contrastive learning the labels are not given.  When labels are not provided, we treat an augmented sample as the positive example.  For a given sample in the batch, the loss function is defined as:
\begin{equation}
    L = - \log \frac{\exp\bigl(\text{sim}(z_{i}, z_{j}) \,/\, \tau\bigr)}
    {\sum_{k \in A(i)} \exp\bigl(\text{sim}(z_{i}, z_{k}) \,/\, \tau\bigr)},
\end{equation}
where $z_{i}$ is the anchor representation, $z_{j}$ is its positive counterpart, 
$\text{sim}(\cdot,\cdot)$ is a similarity measure (e.g., cosine similarity), 
$\tau$ is a temperature parameter, and $A(i)$ is the set of all possible 
contrasting samples (including positives and negatives) for the anchor $z_{i}$.
    
\end{itemize}

\subsection{\textbf{Frame Aggregation (Video-to-Image Transformation):}} Typically, video-based methods process frames either sequentially (e.g., via LSTMs) or by incorporating time-dimension convolutions (e.g., 3D CNNs). In contrast, we propose a \textit{frame aggregation} technique that spatially arranges multiple frames into a single image, allowing us to leverage standard 2D backbones.

Let $V^{a}$ be a video with $T$ frames, and let $\{v_{1}, v_{2}, \dots, v_{y}\}$ be a subset of $y \leq T$ frames sampled from $V^{a}$. Each frame $v_i$ is resized to a canonical resolution, $\mathbf{X}_{i} \in \mathbb{R}^{h \times w \times 3}$. We then assemble these $x$ frames into a grid of size $n \times m$ (with $n \times m \geq y$). Concretely, define an empty canvas
$\mathbf{Z} \in \mathbb{R}^{(n \cdot h) \times (m \cdot w) \times 3},$ where each element $\mathbf{Z}[u, v]$ is initialized to zero (or an equivalent blank color). For each frame $v_{i}$, we assign a grid location from left to right and top to bottom. Once the process is repeated for all selected frames, $\mathbf{Z}$ becomes our \emph{aggregated image}. So, to aggregate 16 frames into a $(224 \times 224 \times 3)$ image we resize each frame into $(56 \times 56 \times 3)$ image and then copy the pixels into a blank image. The frames are arranged sequentially from left to right and from top to bottom. Figure~\ref{fig:version_comparison} conceptually illustrates how frames are placed in a 2D layout.

\subsection{\textbf{Dual-Input Architecture}}
\label{sec:dual_input_arch}

Conventional contrastive frameworks typically process a single input and its augmentations across separate passes. In contrast, we adopt a \emph{dual-input} design that ingests two aggregated video-images in parallel. Let
\(
\mathbf{x}_1, \mathbf{x}_2 \in \mathbb{R}^{H \times W \times 3}
\)
be these two inputs. A shared encoder network maps them to feature vectors
\(
\mathbf{f}_1 \text{ and } \mathbf{f}_2,
\)
which are then projected to 
\(
\mathbf{z}_1 \text{ and } \mathbf{z}_2.
\)
These projected embeddings are concatenated for our supervised contrastive loss, where label information guides positive and negative pair assignments. 

\paragraph{Advantages.} 
\textbf{(1)}~Processing two parallel inputs under the same label augments the set of positive examples, promoting stronger class-consistent embeddings.  
\textbf{(2)}~Reusing a single backbone for both inputs incurs minimal additional overhead, making the approach computationally efficient for learning large-scale video representation.

\subsection{\textbf{Supervised Contrastive Objective}}

In standard supervised contrastive learning \cite{khosla2020supervised}, positive pairs are typically constructed using an image and its augmented version, ensuring that the learned representations remain invariant to transformations. However, in our framework, we leverage the temporal nature of video data by defining positive pairs using two different frame aggregations of the same video, alongside other videos belonging to the same class label. This approach enhances intraclass alignment while capturing meaningful variations within video sequences.

Formally, for a batch of $N$ original video samples, each sample is transformed into two aggregated frame representations, leading to a total of $2N$ embeddings. Given an embedding $z_i$ associated with a video, supervised contrastive loss is defined as:

\begin{equation}
    \mathcal{L}_{\text{SCFA}} 
    = 
    - \frac{1}{2N}
    \sum_{i=1}^{2N}
    \log \Biggl(
      \frac{
        \sum_{j \in P(i)} 
          \exp\bigl( S_{ij} \bigr)
      }{
        \sum_{k \neq i}^{2N} 
          \exp\bigl( S_{ik} \bigr)
      }
    \Biggr),
\end{equation}

where $P(i)$ represents the set of indices corresponding to positive samples, which includes embeddings derived from the same video, as well as embeddings of videos that share the same class label. The similarity score is computed as:

\begin{equation}
    S_{ij} = \frac{z_i^{\top} z_j}{\tau},
\end{equation}

where $\tau$ is a temperature scaling parameter that controls the sharpness of the similarity distribution.

By requiring that each video sample appear twice in the batch with different frame aggregations, our approach ensures that positive pairs incorporate temporal variations while maintaining class consistency. Unlike traditional supervised contrastive frameworks that primarily rely on augmentations, our method directly exploits the intrinsic temporal structure of video data, leading to more robust and discriminative representations. Furthermore, by employing a dual input architecture inspired by the Barlow twins framework, we optimize the contrastive loss across both spatial and temporal domains efficiently.

\section{\textbf{Experiments}}
\label{Experiments}

We implement a supervised contrastive learning framework using a ResNet50 backbone for feature extraction and a projection head. A contrastive loss function enforces positive pair selection via a temperature-scaled similarity measure with a mask-based approach. A data generator aggregates video frames into an image and constructs augmented views for contrastive training\cite{chowdhury2024active}. Different views are generated from random sampling of video frames. The encoder model is optimized using Adam optimizer ($1e-4$ learning rate) with a batch size of 32 over 20 epochs. To evaluate the learned representations, we build a classifier using the encoder model. This classifier accepts aggregated video data in the same way as before. The backbone of the encoder model is fine-tuned using a soft-max loss with categorical labels.

We use ResNet-50 as the backbone of the encoder for our SCFA framework. For each video, 16 frames are sampled and spatially arranged to form a aggregated input $224 \times 224 \times 3$. These frames are selected by random temporal sampling during training.

The projection head consists of a two-layer MLP with ReLU activation in between, followed by L2 normalization. We used a projection size of 128. The model is trained using a batch size of 64 for 100 epochs using the Adam optimizer. The initial learning rate is set to 0.001 with cosine annealing for learning rate scheduling. We used a temperature parameter of 0.07 for contrastive loss.

We conduct experiments on two standard video action recognition datasets:

\textbf{HMDB51} is a diverse benchmark data set that contains 6,766 video clips in 51 categories of human action, sourced from movies and online videos \cite{kuehne2011hmdb}.

\textbf{Penn Action} comprises 2,326 video sequences annotated with 15 body keypoints in 15 action classes, focused on fine-grained human pose and motion analysis \cite{luvizon20182d}.

\subsection{\textbf{Baseline Comparisons}}

We compare our SCFA method with the following baselines listed in Table~\ref{tab:penn_results}:

\begin{itemize}

    \item \textbf{ViVIT}: A video transformer model that processes 16 randomly selected consecutive frames using a BERT-style attention mechanism to learn spatio-temporal representations \cite{arnab2021vivit}.

    \item  \textbf{Timesformer}: A transformer-based video classifier (TimeSformer\cite{bertasius2021space}) uses divided spatiotemporal self-attention to achieve strong action recognition performance with faster training and scalability for long clips.
    
    \item \textbf{ResNet with Aggregation}: A supervised ResNet-50 model trained from scratch using our video frame aggregation method, but without any contrastive learning. This model is trained using softmax loss for classification.
    
    \item \textbf{VIT with Aggregation}: An ImageNet-21k pre-trained Vision Transformer applied to our aggregated image input and fine-tuned with softmax loss \cite{dosovitskiy2020image}.
    
\end{itemize}

These comparisons should demonstrate the effectiveness of SCFA over both traditional image-based backbones and video transformer models.

\begin{table}[h]
    \centering
    \caption{Comparison of classification accuracy (\textbf{mean ± std}) on the Penn Action dataset. SCFA significantly outperforms baseline models while using the same input size as image-based backbones.}

    \begin{tabular}{lcc}
        \toprule
        Method & Input Size & Accuracy \\
        \midrule
        ViVIT & (16,48,48,3) & 43.21 ± 0.62\% \\
        Timesformer & (16,48,48,3) & 49.52 ± 0.52\% \\
        ResNet with Aggregation & (224,224,3) & 46.36 ± 1.03\% \\
        VIT with Aggregation & (224,224,3) & 48.87 ± 2.04\% \\
        SCFA (Ours) & (224,224,3) & \textbf{76.74 ± 0.39\%} \\
        \bottomrule
    \end{tabular}
    \label{tab:penn_results}
\end{table}

\begin{table}[h]
    \centering
    \caption{Classification accuracy (\textbf{mean ± std}) on the HMDB dataset. Our SCFA approach achieves the best performance among all tested methods, demonstrating strong generalization.}

    \begin{tabular}{lcc}
        \toprule
        Method & Input Size & Accuracy \\
        \midrule
        ViVIT & (16,48,48,3) & 36.25 ± 0.70\% \\
        Timesformer & (16,48,48,3) & 41.04 ± 0.22\% \\
        ResNet with Aggregation & (224,224,3) & 25.03 ± 0.71\% \\
        VIT with Aggregation & (224,224,3) & 40.25 ± 1.40\% \\
        SCFA (Ours) & (224,224,3) & \textbf{48.37 ± 0.43\%} \\
        \bottomrule
    \end{tabular}
    \label{tab:hmdb_results}
\end{table}

We evaluated our method on the Penn Action and HMDB datasets, comparing it with existing video representation learning approaches. The results demonstrate that our Supervised Contrastive Frame Aggregation (SCFA) significantly outperforms prior methods while maintaining computational efficiency. SCFA achieves the highest accuracy across both datasets, highlighting its effectiveness in learning video representations while being computationally efficient.

The experimental results clearly highlight the strength of our SCFA framework. In both the Penn Action and HMDB datasets, SCFA achieves substantial improvements over baseline methods, demonstrating a gain of over 27\% in Penn Action and more than 8\% on HMDB compared to the next-best method. These improvements are particularly noteworthy given that our approach uses the same input size as image-based models and avoids the computational burden of 3D convolutions or full attention over spatio-temporal sequences.

This method can occasionally lead to the omission of unique frames. However, since frame sampling is randomized in each training iteration, the probability that a specific frame $f_i$ is never sampled across $B$ batches is given by: $P(f_i \notin \text{batch}) = \left(1 - \frac{1}{T}\right)^{By},$
where $T$ is the total number of frames in the video and $y$ is the number of frames sampled per batch. For large batch size $B$, this probability approaches zero, ensuring nearly complete coverage of frames throughout training. Thus, while some important frames may be missed in individual samples, the self-supervised framework compensates over time with diverse views and repeated exposure.

\subsection*{\textbf{Conclusion}}

In this paper, we introduced the \textbf{Supervised Contrastive Frame Aggregation (SCFA)} framework for efficient video representation learning. SCFA uses a novel video-to-image aggregation strategy, enabling the use of image-based CNNs like ResNet-50 without expensive 3D or transformer-based architectures. The results of the Penn Action dataset demonstrate a significant increase in performance, achieving \textbf{ 76. 74\% accuracy}, well above previous baselines. The approach also generalizes well to HMDB, confirming the robustness of the method. Our source code is available on \href{https://anonymous.4open.science/r/SCFA-04D4/}{https://anonymous.4open.science/r/SCFA-04D4/}.

Our results underscore the effectiveness of the frame aggregation strategy, which enables the reuse of powerful 2D CNN backbones by converting temporal video information into spatial patterns. Combined with the supervised contrastive loss and dual-input setup, the framework efficiently captures both intra-class variability and temporal consistency. This design not only improves the representation learning for video data, but also offers a scalable and resource-efficient alternative to traditional video models, making it well suited for downstream tasks in real-world scenarios.

\end{document}